\documentclass[11pt,a4paper]{article}
\usepackage[hyperref]{emnlp2020}
\usepackage{times}
\usepackage{latexsym}

\usepackage{xcolor}
\usepackage{makecell}
\usepackage{graphicx}
\usepackage{tikz}
\usepackage{multirow}
\usepackage{booktabs,tabularx,enumitem,ragged2e}
\usepackage{url}

\usepackage{microtype}
\usepackage[normalem]{ulem} 

\aclfinalcopy

\title{Adversarial Augmentation Policy Search for Domain and Cross-Lingual Generalization in Reading Comprehension}

\author{Adyasha Maharana \,\,\,\,\,\,\,\,\,\,\,\,\,\,\,\,\,\,\,\,\, Mohit Bansal \\
  UNC Chapel Hill \\
  {\tt \{adyasha, mbansal\}@cs.unc.edu} \\
}

\date{}

\begin{document}
\maketitle
\begin{abstract}
Reading comprehension models often overfit to nuances of training datasets and fail at adversarial evaluation. Training with adversarially augmented dataset improves robustness against those adversarial attacks but hurts generalization of the models. In this work, we present several effective adversaries and automated data augmentation policy search methods with the goal of making reading comprehension models more robust to adversarial evaluation, but also improving generalization to the source domain as well as new domains and languages. We first propose three new methods for generating QA adversaries, that introduce multiple points of confusion within the context, show dependence on insertion location of the distractor, and reveal the compounding effect of mixing adversarial strategies with syntactic and semantic paraphrasing methods. Next, we find that augmenting the training datasets with uniformly sampled adversaries improves robustness to the adversarial attacks but leads to decline in performance on the original unaugmented dataset. We address this issue via RL and more efficient Bayesian policy search methods for automatically learning the best augmentation policy combinations of the transformation probability for each adversary in a large search space. Using these learned policies, we show that adversarial training can lead to significant improvements in in-domain, out-of-domain, and cross-lingual (German, Russian, Turkish) generalization.\footnote{We will publicly release all our code, adversarial policy data, and models on our webpage.}

\end{abstract}

\section{Introduction}
There has been growing interest in understanding NLP systems and exposing their vulnerabilities through maliciously designed inputs \cite{iyyer2018adversarial, belinkov2017synthetic, nie2019analyzing, gurevych2018proceedings}. Adversarial examples are generated using search \cite{alzantot2018generating}, heuristics \cite{jia2017adversarial} or gradient \cite{ebrahimi2017hotflip} based techniques to fool the model into giving the wrong outputs. Often, the model is further trained on those adversarial examples to make it robust to similar attacks. In the domain of reading comprehension (RC), adversaries are QA samples with distractor sentences that have significant overlap with the question and are randomly inserted into the context. By having a fixed template for creating the distractors and training on them, the model identifies learnable biases and overfits to the template instead of being robust to the attack itself \cite{jia2017adversarial}. 
Hence, we first build on \citet{wang2018robust}'s work of adding randomness to the template and significantly expand the pool of distractor candidates by introducing multiple points of confusion within the context, adding dependence on insertion location of the distractor, and further combining distractors with syntactic and semantic paraphrases to create combinatorially adversarial examples that stress-test the model's language understanding capabilities. These adversaries inflict up to 45\% drop in performance of RC models built on top of large pretrained models like RoBERTa \cite{liu2019roberta}.

Next, to improve robustness to the aforementioned adversaries, we finetune the RC model with a combined augmented dataset containing an equal number of samples from all of the adversarial transformations. While it improves robustness by a significant margin, it leads to decline in performance on the original unaugmented dataset. Hence, instead of uniformly sampling from the various adversarial transformations, we propose to perform a search for the best adversarial policy combinations that improve robustness against the adversarial attacks and also preserve/improve accuracy on the original dataset via data augmentation. However, it is slow, expensive and inductive-biased to manually tune the transformation probability for each adversary and repeat the process for each target dataset, and so we present RL and Bayesian search methods to learn this policy combination automatically. 

For this, we create a large augmentation search space of up to 10\textsuperscript{6}, with four adversarial methods, two paraphrasing methods and a discrete binning of probability space for each method (see Figure \ref{fig:flowchart}). \citet{cubuk2018autoaugment} showed via AutoAugment that a RNN controller can be trained using reinforcement learning to find the best policy in a large search space. However, AutoAugment is computationally expensive and relies on the assumption that the policy searched using rewards from a smaller model and reduced dataset will generalize to bigger models. Alternatively, the augmentation methods can be modelled with a surrogate function, such as Gaussian processes \cite{rasmussen2003gaussian}, and subjected to Bayesian optimization~\cite{snoek2012practical}, drastically reducing the number of training iterations required for achieving similar results (available as a software package for computer vision).\footnote{\url{https://pypi.org/project/deepaugment/}} Hence, we extend these ideas to NLP and perform a systematic comparison between AutoAugment and our more efficient BayesAugment.

Finally, there has been limited previous work exploring the role of adversarial data augmentation to improve generalization of RC models to out-of-domain and cross-lingual data. Hence, we also perform automated policy search of adversarial transformation combinations for enhancing generalization from English Wikipedia to datasets in other domains (news, web) and languages (Russian, German, Turkish). Policy search methods like BayesAugment can be readily adapted for low-resource scenarios where one only has access to a small development set that the model can use as a black-box evaluation function (for rewards, but full training or gradient access on that data is unavailable). We show that augmentation policies for the source domain learned using target domain performance as reward, improves the model's generalization to the target domain with only the use of a small development set from that domain. Similarly, we use adversarial examples in a pivot language (in our case, English) to improve performance on other languages' RC datasets using rewards from a small development set from that language. 

Our contributions can be summarized as follows:
\begin{itemize}[nosep, wide=0pt, leftmargin=*, after=\strut]
    \item We first propose novel adversaries for reading comprehension that cause up to 45\% drop in large pretrained models' performance. Augmenting the training datasets with uniformly sampled adversaries improves robustness to the adversarial attacks but leads to decline in performance on the original unaugmented dataset. 
    \item We next demonstrate that optimal adversarial policy combinations of transformation probabilities (for augmentation and generalization) can be automatically learned using policy search methods. Our experiments show that efficient Bayesian optimization achieves similar results as AutoAugment with a fraction of the resources.
    \item By training on the augmented data generated via the learned policies, we not only improve adversarial robustness of the models but also show significant gains i.e., up to 2.07\%, 5.0\%, and 2.21\% improvement for in-domain, out-of-domain, and cross-lingual evaluation respectively.
    
Overall, the goal of our paper is to make reading comprehension models robust to adversarial attacks as well as out-of-distribution data in cross-domain and cross-lingual scenarios.

\end{itemize}

\begin{table*}[t]
\small
\centering
\def\arraystretch{1.5}
\begin{tabular}{|p{2.6cm}|p{3.5cm}|p{8.5cm}|}
\hline
  \textbf{Adversary Method} & \textbf{Description} & \textbf{Original Question/Sentence and Corresponding Distractor} \\
  \hline
  AddSentDiverse & \cite{jia2017adversarial, wang2018robust} & {Q: In what country is Normandy located? \newline D: \textit{D-Day} is located in the country of \textit{Sri Lanka.}} \\
  \hline
  AddKSentDiverse & Multiple AddSentDiverse distractors are inserted randomly in the context. & {Q: Which county is developing its business center? \newline D1: The county of Switzerland is developing its art periphery. \newline D2: The county of Switzerland is developing its \textit{home center}.} \\
  \hline
  AddAnswerPosition & Answer span is preserved in this distractor. It is most misleading when inserted before the original answer. & {Q: What is the steam engine's thermodynamic basis? \newline A: The Rankine cycle is the fundamental thermodynamic underpinning of the steam engine. \newline D: Rankine cycle is the \textit{air} engine's thermodynamic basis.} \\
  \hline
  InvalidateAnswer & AddSentDiverse and additional elimination of the original answer. & {Q: Where has the official home of the Scottish Parliament been since 2004? \newline D: Since \textit{October 2002}, the \textit{unofficial abroad} of the \textit{Welsh Assembly} has been a \textit{old Welsh Assembly Houses}, in the \textit{Golden Gate Bridge} area of \textit{Glasgow}.} \\
  \hline
  PerturbAnswer & Content words (except named entities) are algorithmically replaced with synonyms and evaluated for consistency using language model. & {A: The UK refused to sign the Social Charter and was exempt from the legislation covering Social Charter issues unless it agreed to be bound by the legislation. \newline P: The UK \textit{repudiated} to \textit{signature} the Social Charter and was exempt from the legislation \textit{encompassing} Social Charter issues unless it \textit{consented} to be \textit{related} by the legislation.} \\
  \hline
  PerturbQuestion & Syntacting paraphrasing network is used to generate the source question with a different syntax. & {Q: In what country is Normandy located? \newline P: \textit{Where} does Normany \textit{exist}?}\\
  \hline
\end{tabular}
\vspace{-5pt}
\caption{Demonstration of the various adversary functions used in our experiments (Q=Question, D=Distractor, A=Answer, P=Paraphrase). Words that have been modified using adversarial methods are italicized in the distractor. \label{tab:adversaryDemo}\vspace{-20pt}}
\end{table*}

\vspace{-10pt}

\section{Related Work}

\textbf{Adversarial Methods in NLP:} Following the introduction of adversarial evaluation for RC models by \citet{jia2017adversarial, wang2018robust}, several methods have been developed for probing the sensitivity and stability of NLP models \cite{nie2019analyzing, glockner2018breaking}. \citet{zhao2017generating} employ GANS to generate semantically meaningful adversaries. \citet{ren2019generating} and \citet{alzantot2018generating} use a synonym-substitution strategy while \citet{ebrahimi2017hotflip} create gradient-based perturbations. \citet{iyyer2018adversarial} construct a syntactic paraphrasing network to introduce syntactic variance in adversaries.

 \textbf{Augmentation and Generalization:} \citet{goodfellow2014explaining} and \citet{miyato2018virtual} use adversarial training to demonstrate improvement in image recognition. \citet{xie2019adversarial} improve the adversarial training scheme with auxiliary batch normalization modules. Back-translation \cite{yu2018qanet}, pre-training with other QA datasets \cite{devlin2018bert, lewis-etal-2019-unsupervised, talmor2019multiqa, longpre2019exploration} and virtual adversarial training \cite{miyato2016adversarial, yang2019improving} are shown to be effective augmentation techniques for RC datasets. \citet{cao2019unsupervised} propose a conditional adversarial self-training method to reduce domain distribution discrepancy. \citet{lee2019domain, wang2019adversarial} use a discriminator to enforce domain-invariant representation learning \cite{fisch2019mrqa}; \citet{chen2018adversarial} and \citet{zhang2017adversarial} learn language-invariant representations for cross-lingual tasks. We show that heuristics-based adversaries can be used for augmentation as well as generalization.

\textbf{Policy Search:} \citet{cubuk2018autoaugment} present the AutoAugment algorithm which uses reinforcement learning to find the best augmentation policies in a large search space, and then follow-up with RandAugment \cite{cubuk2019randaugment} which reduces the task to simple grid-search. \citet{niu2019automatically} use AutoAugment to discover perturbation policies for dialogue generation. \citet{ho2019population} use population-based augmentation (PBA) techniques \cite{jaderberg2017population} and significantly reduce the compute time required by AutoAugment. We are the first to adapt RandAugment style techniques for NLP via our BayesAugment method. RandAugment enforces uniform transformation probability on all augmentation methods and collapses the augmentation policy search space to two global parameters. BayesAugment eliminates the need to choose between adversarial methods and optimizes only for their transformation probabilities (see Sec.~\ref{sec:SearchSpace}).

\section{Adversary Policy Design}
As shown by \citet{jia2017adversarial}, QA models are susceptible to random, semantically meaningless and minor changes in the data distribution. We extend this work and propose adversaries that exploit the model's sensitivity to insert location of distractor, number of distractors, combinatorial adversaries etc. After exposing the model's weaknesses, we strengthen them by training on these adversaries and show that the model's robustness to adversarial attacks significantly increases due to it. Finally, in Sec.~\ref{sec:AutoPolicySearch}, we automatically learn the right combination of transformation probability for each adversary in response to a target improvement using policy search methods.

\subsection{Adversary Transformations}
We present two types of adversaries, namely positive perturbations and negative perturbations (or attacks) (Figure~\ref{fig:flowchart}). Positive perturbations are adversaries generated using methods that have been traditionally used for data augmentation in NLP i.e., semantic and syntactic transformations. Negative perturbations are distractor sentences based on the classic AddSent model \cite{jia2017adversarial} that exploits the RC model's shallow language understanding to mislead it to incorrect answers. We use the method outlined by \citet{wang2018robust} for \textbf{AddSentDiverse} to generate a distractor sentence (see Table~\ref{tab:adversaryDemo}) and insert it randomly within the context of a QA sample.

We introduce more variance to adversaries with \textbf{AddKSentDiverse}, wherein multiple distractor sentences are generated using AddSentDiverse and are inserted at independently sampled random positions within the context. For \textbf{AddAnswerPosition}, the original answer span is retained within the distractor sentence and the model is penalized for incorrect answer span location. We remove the sentence containing the answer span from the context and introduce a distractor sentence to create \textbf{InvalidateAnswer} adversarial samples which are no longer answerable. \textbf{PerturbAnswer} adversaries are created by following the \texttt{Perturb} subroutine \cite{alzantot2018generating} and generating semantic paraphrases of the sentence containing the answer span. We use the syntactic paraphrase network \cite{iyyer2018adversarial} to create \textbf{PerturbQuestion} adversarial samples by replacing the original question with its paraphrase.

Finally, we combine negative and positive perturbations to create adversaries which double-down on the model's language understanding. It always leads to a larger drop in performance when tested on the RC models trained on original unaugmented datasets. See Appendix for more details.

\subsection{Adversarial Policy \& Search Space}
\label{sec:SearchSpace}
Reading comprehension models are often trained with adversarial samples in order to improve robustness to the corresponding adversarial attack. We seek to find the best combination of adversaries for data augmentation that also preserves/improves accuracy on source domain and improves generalization to a different domain or language. 

\begin{figure}[t]
\centering
\resizebox{0.45\textwidth}{!}{
        \includegraphics[width=0.45\textwidth]{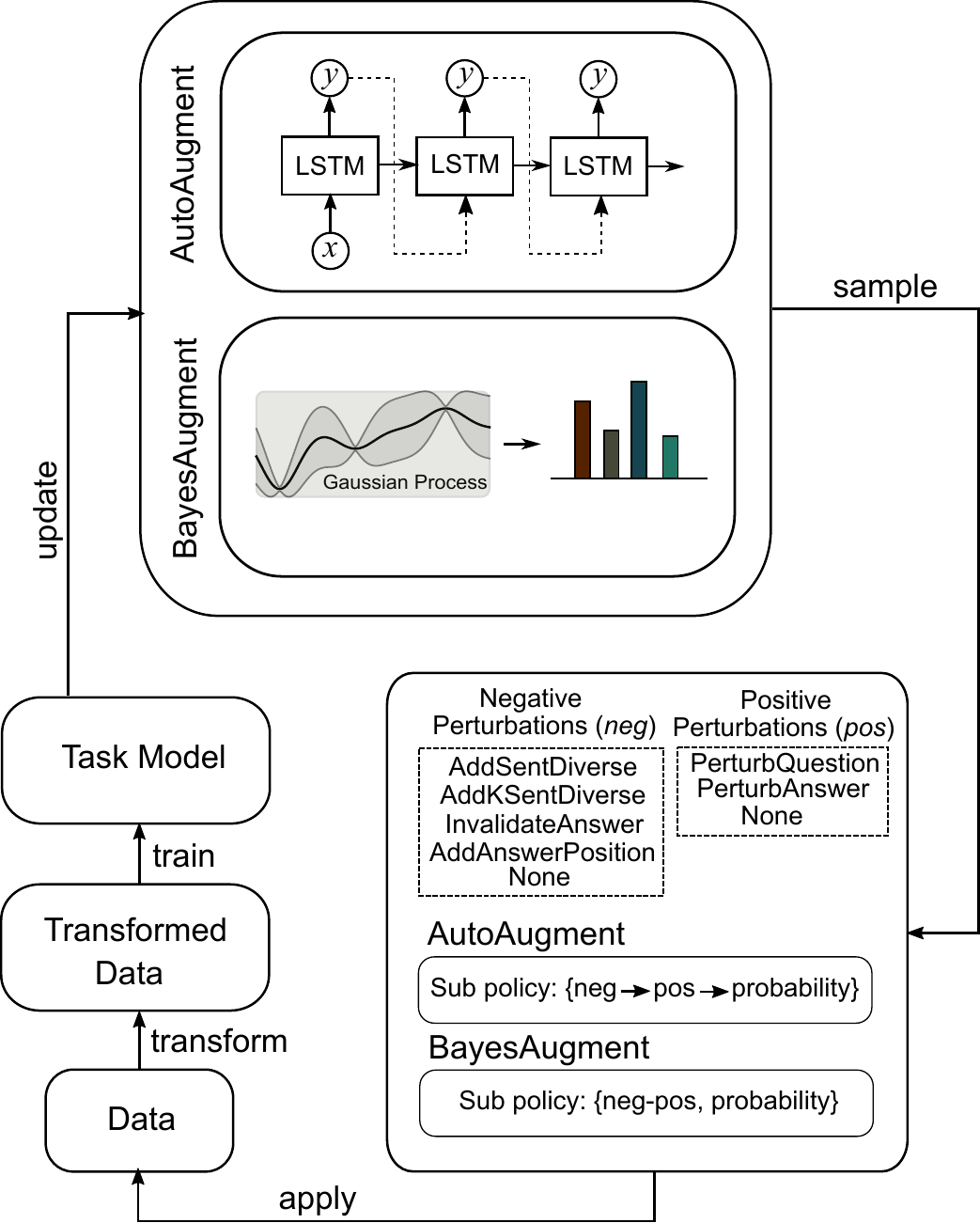}
        }
    \vspace{-5pt}
    \caption{Flow chart of training loop for AutoAugment controller and Bayesian optimizer. See Sec.~\ref{sec:AutoPolicySearch}. \label{fig:flowchart}\vspace{-10pt}}
\end{figure}

\textbf{AutoAugment:} Following previous work in AutoAugment policy search \cite{cubuk2018autoaugment, niu2019automatically}, we define a sub-policy to be a set of adversarial transformations which are applied to a QA sample to generate an adversarial sample. We show that adversaries are most effective when positive and negative perturbations are applied together (Table \ref{tab:advEvaluation}). Hence, to prepare one sub-policy, we select one of the four negative perturbations (or none), combine it with one of the two positive perturbations (or none) and assign the combination a transformation probability (see Figure \ref{fig:flowchart}). The probability space $[0,1]$ is discretized into 6 equally spaced bins. This leads to a search space of $5*3*6=90$ for a single sub-policy. Next, we define a complete adversarial policy as a set of $n$ sub-policies with a search space of $90^n$. For each input QA sample, one of the sub-policies is randomly sampled and applied (with a probability equal to the transformation probability) to generate the adversarial sample. Thus, each original QA sample ends up with one corresponding adversarial sample or none.

\textbf{BayesAugment:} We adopt a simplified formulation of the policy for our BayesAugment method, following \citet{ho2019population} and RandAugment \cite{cubuk2019randaugment}. Sampling of positive and negative adversaries is eliminated and transformation probabilities of all possible combinations of adversaries are optimized over a continuous range $[0, 1]$.\footnote{RandAugment collapses a large parameter space by enforcing uniform probability on all transformations and optimizing for: (i) global distortion parameter, (ii) number of transformations applied to each image. It uses hyperparameter optimization and shows results with naive grid search due to small search space. RandAugment is not directly applicable to our setting because there is no notion of global distortion for text. Hence, we borrow the idea of treating augmentation policy parameters as hyperparameters but use Bayesian optimization for search.} Consequently, one of these combinations is randomly sampled for each input QA sample to generate adversaries. Empirically, the dominant adversary in a policy is the attack with highest transformation probability (see policies in Table \ref{tab:sup_policies} in Appendix). Due to the probabilistic nature of the policy, it is possible for the model to not add any adversarial sample at all, but the probability of this happening is relatively low.\vspace{-5pt}

\section{Automatic Policy Search}
\label{sec:AutoPolicySearch}
Next, we need to perform search over the large space of augmentation policies in order to find the best policy for a desired outcome. Performing naive search (random or grid) or manually tuning the transformation probabilities is slow, expensive and largely impractical due to resource constraints. Hence, we compare two different approaches for learning the best augmentation policy in fewer searches: AutoAugment and BayesAugment. We follow the optimization procedure as demonstrated in Figure \ref{fig:flowchart}. For $t=1,2,...$, do:
\begin{itemize}[nosep, wide=0pt, leftmargin=*, after=\strut]
    \item Sample the next policy $p_{t}$ (\textit{sample})
    \item Transform training data with $p_{t}$ and generate augmented data (\textit{apply, transform})
    \item Train the downstream task model with augmented data (\textit{train})
    \item Obtain score on validation dataset as reward $r_{t}$
    \item Update Gaussian Process or RNN Controller with $r_{t}$ (\textit{update})
\end{itemize}

\subsection{AutoAugment}
Our AutoAugment model (see Figure \ref{fig:flowchart}) consists of a recurrent neural network-based controller and a downstream task model. The controller has $n$ output blocks for $n$ sub-policies; each output block generates distributions for the three components of sub-policies i.e., \textit{neg}, \textit{pos} and \textit{probability}. The adversarial policy is generated by sampling from these distributions and applied on input dataset to create adversarial samples, which are added to the original dataset to create an augmented dataset. The downstream model is trained on the augmented dataset till convergence and evaluated on a given metric, which is then fed back to the controller as a reward (see the \textit{update} flow in figure). We use REINFORCE \cite{sutton2000policy, williams1992simple} to train the controller.

\subsection{BayesAugment}
Typically, it takes thousands of steps to train an AutoAugment controller using reinforcement learning which prohibits the use of large pretrained models as task model in the training loop. For example, the controllers in \citet{cubuk2018autoaugment} were trained for 15,000 samples or more. To circumvent this computational issue, we frame our adversarial policy search as a hyperparameter optimization problem and use Bayesian methods to perform the search. Bayesian optimization techniques use a surrogate model to approximate the objective function $f$ and an acquisition function to sample points from areas where improvement over current result is most likely. The prior belief about $f$ is updated with samples drawn from $f$ in order to get a better estimate of the posterior that approximates $f$. Bayesian methods attempt to find global maximum in the minimum number of steps.

\subsection{Rewards} The F1 score of downstream task model on development set is used as reward during policy search. To discover augmentation policies which are geared towards improving generalization of RC model, we calculate the F1 score of task model (trained on source domain) on out-of-domain or cross-lingual development datasets, and feed it as the reward to the optimizer.

\subsection{Datasets}

We use SQuAD v2.0 \cite{rajpurkar2018know} and NewsQA \cite{trischler2016newsqa} for adversarial evaluation and in-domain policy-search experiments. Futher, we measure generalization from SQuAD v2.0 to NewsQA and TriviaQA \cite{joshi2017triviaqa}, and from SQuAD v1.1 \cite{rajpurkar2016squad} to German dataset from MLQA \cite{lewis2019mlqa} and Russian, Turkish datasets from XQuAD \cite{artetxe2019cross}.\footnote{The choice of cross-lingual datasets in our experiments is based on availability of \textit{x}-en translation and span alignment models for the Translate-Test method~\cite{asai2018multilingual}} See Appendix for more details on datasets and training.
\vspace{-5pt}

\subsection{Reading Comprehension Models} We use RoBERTa\textsubscript{BASE} as the primary RC model for all our experiments. For fair baseline evaluation on out-of-domain and cross-lingual datasets, we also use the development set of the target task to select the best checkpoint. Search algorithms like AutoAugment require a downstream model that can be trained and evaluated fast, in order to reduce training time. So, we use distilRoBERTa\textsubscript{BASE} \cite{sanh2019distilbert} for AutoAugment training loops. BayesAugment is trained for fewer iterations than AutoAugment and hence, allows us to use RoBERTa\textsubscript{BASE} model directly in the training loop. See Appendix for more details and baseline performances of these models.
\vspace{-5pt}

\subsection{Evaluation Metrics}
We use the official SQuAD evaluation script for evaluation of robustness to adversarial attacks and performance on in-domain and out-of-domain datasets.\footnote{Statistical significance is computed with 100K samples using bootstrap \cite{noreen1989computer, tibshirani1993introduction}.} For cross-lingual evaluation, we use the modified Translate-Test method as outlined in \citet{lewis2019mlqa, asai2018multilingual}. QA samples in languages other than English are first translated to English and sent as input to RoBERTa\textsubscript{BASE} finetuned on SQuAD v1.1. The predicted answer spans within English context are then mapped back to the context in original language using alignment scores from the translation model. We use the top-ranked German$\rightarrow$English and Russian$\rightarrow$English models in WMT19 shared news translation task, and train a Turkish$\rightarrow$English model using a similar architecture, to generate translations and alignment scores \cite{ng2019facebook}.\footnote{\url{https://github.com/pytorch/fairseq}}

\begin{table}[t]
\small
\begin{center}
\def\arraystretch{1.3}
\resizebox{0.42\textwidth}{!}{
\begin{tabular}{|l|ll|}
 \hline
 \textbf{Adversary Method} & \textbf{SQuAD} & \textbf{NewsQA} \\
 \hline
 Baseline (No Adversaries) & 81.17 & 58.40 \\
 AddSentDiverse & 65.50 & 51.47 \\ 
 AddKSentDiverse (K=2) & 45.31 & 48.31 \\ 
 AddAnswerPosition & 68.91 & 49.20 \\
 InvalidateAnswer & 77.75 & 24.03 \\
 PerturbQuestion & 43.67 & 36.76 \\
 PerturbAnswer & 71.97 & 59.08 \\
 \hline
 \multicolumn{3}{|c|}{\textit{Effect of Multiple Distractors}} \\
 \hline
 AddSentDiverse & 65.50 & 51.47 \\ 
 Add2SentDiverse & 45.31 & 48.31 \\ 
 Add3SentDiverse & 43.49 & 44.81\\ 
 \hline
 \multicolumn{3}{|c|}{\textit{Combinatorial effect}}\\
 \hline
 AddSentDiverse & 65.50 & 51.47\\ 
 {\hspace{0.5cm} + PerturbAnswer} & 50.71 & 51.43\\ 
 AddKSentDiverse & 45.31 & 48.31 \\
 {\hspace{0.5cm} + PerturbQuestion} & 31.56 & 29.56 \\
 \hline
 \multicolumn{3}{|c|}{\textit{Effect of Insert Location of AddAnswerPosition}}\\
 \hline
 Random & 68.91 & 49.20 \\ 
 Prepend & 66.52 & 48.01 \\ 
 Append & 67.84 & 48.76 \\ 
 \hline
\end{tabular}}
\vspace{-5pt}
\caption{Adversarial evaluation of baseline RoBERTa\textsubscript{BASE} trained on SQuAD v2.0 and NewsQA. Compare to corresponding rows in Table~\ref{tab:afterAdvTraining} to observe difference in performance after adversarial training. Results (F1 score) are shown on dev set. \label{tab:advEvaluation}\vspace{-15pt}}
\end{center}
\end{table}

\section{Results}
First, in Sec.~\ref{sec:adveval}, we perform adversarial evaluation of baseline RC models for various categories of adversaries. Next, in Sec.~\ref{sec:advtrain}, we train the RC models with an augmented dataset that contains equal ratios of adversarial samples and show that it improves robustness to adversarial attacks but hurts performance of the model on original unaugmented dataset. Finally, in Sec.~\ref{sec:AugPolicyResults}, we present results from AutoAugment and BayesAugment policy search and the in-domain, out-of-domain and cross-lingual performance of RC models trained using augmentation data generated from the learned policies with corresponding target rewards.

\subsection{Adversarial Evaluation}  
\label{sec:adveval}
Table~\ref{tab:advEvaluation} shows results from adversarial evaluation of RoBERTa\textsubscript{BASE} finetuned with SQuAD v2.0 and NewsQA respectively. All adversarial methods lead to a significant drop in performance for the finetuned models i.e., between 4-45\% for both datasets. The decrease in performance is maximum when there are multiple distractors in the context (Add3SentDiverse) or perturbations are combined with one another (AddSentDiverse + PerturbAnswer). These results show that, in spite of being equipped with a broader understanding of language from pretraining, the finetuned RC models are shallow and over-stabilized to textual patterns like n-gram overlap. Further, the models aren't robust to semantic and syntactic variations in text.

Additionally, we performed manual evaluation of 96 randomly selected adversarial samples (16 each from attacks listed in Table \ref{tab:adversaryDemo}) and found that a human annotator picked the right answer for 85.6\% of the questions.

\begin{table}[t!]
\small
\begin{center}
\def\arraystretch{1.3}
\resizebox{0.42\textwidth}{!}{
\begin{tabular}{|l|ll|}
 \hline
 \textbf{Adversary Method} & \textbf{SQuAD} & \textbf{NewsQA} \\
 \hline
 AddSentDiverse & 68.00 & 61.13 \\ 
 AddKSentDiverse (K=2) & 79.44 & 62.31 \\ 
 AddAnswerPosition & 80.16 & 56.90 \\
 InvalidateAnswer & 91.41 & 67.57 \\
 PerturbQuestion & 60.91 & 44.99 \\
 PerturbAnswer & 76.42 & 60.74 \\
 \hline
 Original Dev (No Adversaries) & 78.83 & 58.08 \\
 \hline
\end{tabular}}
\vspace{-5pt}
\caption{Adversarial evaluation after training RoBERTa\textsubscript{BASE} with the original dataset augmented with equally sampled adversarial data. Compare to corresponding rows in Table~\ref{tab:advEvaluation} to observe difference in performance after adversarial training. Results (F1 score) are shown on dev set. \label{tab:afterAdvTraining} \vspace{-20pt}}
\end{center}
\end{table}

\subsection{Manual Adversarial Training}
\label{sec:advtrain}
Next, in order to remediate the drop in performance observed in Table~\ref{tab:advEvaluation} and improve robustness to adversaries, the RC models are further finetuned for 2 epochs with an adversarially augmented training set. The augmented training set contains each QA sample from the original training set and a corresponding adversarial QA sample by randomly sampling from one of the adversary methods. Table~\ref{tab:afterAdvTraining} shows results from adversarial evaluation after adversarial training. Adding perturbed data during training considerably improves robustness of the models to adversarial attacks. For instance, RoBERTa\textsubscript{BASE} performs with 79.44 F1 score on SQuAD AddKSentDiverse samples (second row, Table~\ref{tab:afterAdvTraining}), as compared to 45.31 F1 score without adversarial training (third row, Table~\ref{tab:advEvaluation}). Similarly, RoBERTa\textsubscript{BASE} performs with 44.99 F1 score on NewsQA PerturbQuestion samples (fifth row, Table~\ref{tab:afterAdvTraining}), as compared to a baseline score of 36.76 F1 score (sixth row, Table~\ref{tab:advEvaluation}). However, this manner of adversarial training also leads to drop in performance on the original unaugmented development set, e.g., RoBERTa\textsubscript{BASE} performs with 78.83 and 58.08 F1 scores on the SQuAD and NewsQA development sets respectively, which is 2.34 and 0.32 points lesser than the baseline (first row, Table~\ref{tab:advEvaluation}).

\begin{table}[t]
\small
\begin{center}
\def\arraystretch{1.3}
\resizebox{0.39\textwidth}{!}{
\makebox[0.5\textwidth]{\begin{tabular}{|p{7mm}|p{17mm}p{17mm}|p{17mm}p{17mm}|} 
 \hline
  Search &  \multicolumn{2}{c|}{\textit{In-domain}}  & \multicolumn{2}{c|}{SQuAD $\rightarrow$}\\
 {Method} & {SQuAD} & {NewsQA} & {NewsQA} & {TriviaQA}\\
 \hline
\multicolumn{5}{|c|}{Validation}\\
 \hline
 Base & 81.17 / 77.54 & 58.40 / 47.04 & 48.36 / 36.06 & 41.60 / 34.86\\
 UniS & 78.83 / 74.68 & 58.08 / 46.79 & 48.24 / 36.03 & 42.04 / 35.11 \\
Auto & 81.63 / 78.06 & \textbf{62.17} / \textbf{49.41} & \textbf{50.57} / \textbf{38.56} & 42.41 / 35.41\\
Bayes & \textbf{81.71} / \textbf{78.12} & 58.62 / 47.21 & 49.73 / 38.38 & \textbf{43.96} / \textbf{36.67}\\
\hline
\multicolumn{5}{|c|}{Test}\\
 \hline
 Base & 80.64 / 77.19 & 57.02 / 45.29 & 44.95 / 34.68 & 36.01 / 29.23\\
 UniS & 78.42 / 75.87 & 57.21 / 45.36 & 46.30 / 35.94 & 37.83 / 30.52 \\
 Auto  & \textbf{81.06} / \textbf{77.79} & \textbf{59.09} / \textbf{45.49} & 46.82 / 35.75 & 37.88 / 30.60 \\ 
 Bayes & 80.88 / 77.57 & 57.63 / 45.32 & \textbf{48.95} / \textbf{37.44} & \textbf{40.99} / \textbf{33.68} \\ 
 \hline
\end{tabular}}
}
\vspace{-5pt}
\caption{Baseline results (first row) and evaluation after finetuning baseline models with the adversarial policies derived from AutoAugment and BayesAugment for in-domain improvements and out-of-domain generalization from Wikipedia (SQuAD) to news (NewsQA) and web (TriviaQA) domains. Results (F1 / Exact Match) are shown on validation and test sets. (Base=Baseline, UniS=Uniform Sampling, Auto=AutoAugment, Bayes=BayesAugment)\label{tab:mainDev}\vspace{-15pt}}
\end{center}
\end{table}

\subsection{Augmentation Policy Search for Domain and Language Generalization}\
\label{sec:AugPolicyResults}
Following the conclusion from Sec.~\ref{sec:advtrain} that uniform sampling of adversaries is not the optimal approach for model performance on original unaugmented dataset, we perform automated policy search over a large search space using BayesAugment and AutoAugment for in-domain as well as cross-domain/lingual improvements (as discussed in Sec.~\ref{sec:AutoPolicySearch}). For AutoAugment, we choose the number of sub-policies in a policy to be $n=3$ as a trade-off between search space dimension and optimum results. We search for the best transformation policies for the source domain that lead to improvement of the model in 3 areas: 1. in-domain performance 2. generalization to other domains and 3. generalization to other languages. These results are presented in Tables \ref{tab:mainDev} and \ref{tab:multilingualDev}, adversarial evaluation of the best BayesAugment models is presented in Table~\ref{tab:BayesAdv}, and the learned policies are shown in the Appendix.

\begin{table}[t!]
\small
\begin{center}
\def\arraystretch{1.3}
\resizebox{0.47\textwidth}{!}{
\begin{tabular}{|p{12mm}|p{17mm}p{17mm}p{17mm}|} 
 \hline
   {Search}& \multicolumn{3}{c|}{\textit{Cross-lingual generalization}}\\
  Method & \multicolumn{3}{c|}{\textit{from English SQuAD} $\rightarrow$}\\
 & {MLQA (de)} & {XQuAD (ru)} & {XQuAD (tr)}\\
  \hline
\multicolumn{4}{|c|}{Validation}\\
 \hline
 Baseline & 58.58 / 36.41 & 67.89 / 44.62 & 42.95 / 25.09 \\
 UniformS & 58.97 / 36.68 & 68.11 / 44.84 & 43.12 / 25.26 \\
 BayesAug & \textbf{59.40} / \textbf{37.11} & \textbf{68.73} / \textbf{45.34} & \textbf{44.09} / \textbf{25.73} \\
\hline
\multicolumn{4}{|c|}{Test}\\
 \hline
  Baseline & 57.56 / 36.01 & 60.81 / 33.47 & 40.49 / 23.14 \\
  UniformS & 58.27 / 36.45 & 61.87 / 34.31 & 41.04 / 23.78 \\
BayesAug & \textbf{59.02} / \textbf{38.01} & \textbf{63.03} / \textbf{34.85} & \textbf{41.95} / \textbf{24.17} \\
\hline
\end{tabular}
}
\vspace{-5pt}
\caption{Cross-lingual QA: Translate-Test \cite{lewis2019mlqa} evaluation after finetuning the baseline with adversarial policies derived from BayesAugment for generalization to German (de), Russian (ru) and Turkish (tr) RC datasets. Results (F1 / Exact Match) are shown on validation and test sets. \label{tab:multilingualDev}\vspace{-20pt}}
\end{center}
\end{table}

\textbf{In-domain evaluation: }The best AutoAugment augmentation policies for improving in-domain performance of RoBERTa\textsubscript{BASE} on the development sets result in 0.46\% and 3.77\% improvement in F1 score over baseline for SQuAD v2.0 and NewsQA respectively (see Table \ref{tab:mainDev}). Similarly, we observe 0.54\% (p=0.021) and  0.22\% (p=0.013) absolute improvement in F1 Score for SQuAD and NewsQA respectively by using BayesAugment policies. This trend is reflected in results on the test set as well. AutoAugment policies result in most improvement i.e., 0.42\% (p=0.014) and 2.07\% (p=0.007) for SQuAD and NewsQA respectively. Additionally, both policy search methods outperform finetuning with a dataset of uniformly sampled adversaries (see row 2 in Table \ref{tab:mainDev}).

\textbf{Out-of-domain evaluation:} To evaluate generalization of the RC model from Wikipedia to news articles and web, we train RoBERTa\textsubscript{BASE} on SQuAD and evaluate on NewsQA and TriviaQA respectively. The baseline row in Table~\ref{tab:mainDev} presents results of RoBERTa\textsubscript{BASE} trained on original unaugmented SQuAD and evaluated on NewsQA and TriviaQA. Next, we reiterate results from Table \ref{tab:afterAdvTraining} and show that finetuning with uniformly sampled dataset (see UniS in Table \ref{tab:mainDev}) of adversaries results in drop in performance on the validation sets of SQuAD and NewsQA. By training on adversarially augmented SQuAD with AutoAugment policy, we see 2.21\% and 0.81\% improvements on the development sets of NewsQA (SQuAD$\rightarrow$NewsQA) and TriviaQA (SQuAD$\rightarrow$TriviaQA) respectively. Similarly, BayesAugment provides 1.37\% and 2.36\% improvements over baseline for development sets of TriviaQA and NewsQA, proving as a competitive and less computationally intensive substitute to AutoAugment. BayesAugment outperforms AutoAugment at out-of-domain generalization by providing 4.0\%(p$<$0.001) and 4.98\% jump on test sets for NewsQA and TriviaQA respectively, as compared to 1.87\% improvements with AutoAugment.

Our experiments suggest that AutoAugment finds better policies than BayesAugment for in-domain evaluation. We hypothesize that this might be attributed to a difference in search space between the two policy search methods. AutoAugment is restricted to sampling at most 3 sub-policies while BayesAugment has to simultaneously optimize the transformation probability for ten or more different augmentation methods. A diverse mix of adversaries from the latter is shown to be more beneficial for out-of-domain generalization but results in minor improvements for in-domain performance. Moving ahead, due to better performance for out-of-domain evaluation and more efficient trade-off with computation, we only use BayesAugment for our cross-lingual experiments.

\textbf{Cross-lingual evaluation: } Table \ref{tab:multilingualDev} shows results of RoBERTa\textsubscript{BASE} finetuned with adversarially augmented SQuAD v1.1\footnote{\textit{InvalidateAnswer} adversaries are not used for generalization from SQuADv1.1 because it does not contain the NoAnswer style samples introduced in SQuADv2.0.} and evaluated on RC datasets in non-English languages. The baseline row presents results from RoBERTa\textsubscript{BASE} trained on original unaugmented SQuAD and evaluated on German MLQA(de), Russian XQuAD(ru) and Turkish XQuAD(tr) datasets; F1 scores on the development sets are 58.58, 67.89 and 42.95 respectively. These scores depend on quality of the translation model as well as the RC model. We observe significant improvements on the development as well as test sets by finetuning baseline RC model with adversarial data from English SQuAD. Uniformly sampled adversarial dataset results in 0.71\% (p=0.063), 1.06\% (p=0.037),  and 0.55\% (p=0.18) improvement for test sets of MLQA(de), XQuAD(ru) and XQuAD(tr), respectively. BayesAugment policies outperform uniform sampling and result in 1.47\% (p=0.004), 2.21\% (p=0.007) and 1.46\% (p=0.021) improvement for test sets of MLQA(de), XQuAD(ru) and XQuAD(tr), respectively.

\begin{table*}[t!]
\small
\begin{center}
\def\arraystretch{1.3}
\resizebox{0.8\textwidth}{!}{
\begin{tabular}{|l|ll|lll|}
 \hline
 & \multicolumn{2}{c|}{\textit{Out-of-domain generalization}} & \multicolumn{3}{c|}{\textit{Cross-lingual generalization}}\\
 \textbf{Adversary Method} & \textbf{TriviaQA} & \textbf{NewsQA} & \textbf{MLQA (de)} & \textbf{XQuAD (ru)} & \textbf{XQuAD (tr)} \\
 \hline
 AddSentDiverse & 
67.17 / 65.60 & 66.26 / 64.59 & 63.68 / 61.09 & 65.21 / 64.04 & 65.17 / 63.83 \\ 
 AddKSentDiverse (K=2) & 78.48 / 76.32 & 77.13 / 75.80 & 76.91 / 74.45 & 77.76 / 75.20 & 77.93 / 75.37\\ 
 AddAnswerPosition & 80.05 / 77.41 & 79.46 / 76.31 & 78.62 / 75.59 & 80.24 / 77.38 & 79.51 / 76.28 \\
 InvalidateAnswer & 88.23 / 85.56 & 90.18 / 78.25 & - & - & -\\
 PerturbQuestion & 60.39 / 58.02 & 54.65 / 51.48 & 58.14 / 56.33 & 60.15 / 57.92 & 59.71 / 56.27 \\
 PerturbAnswer & 77.12 / 75.38 & 76.30 / 74.12 & 77.28 / 75.82 & 74.31 / 72.88 & 74.72 / 73.16 \\
 \hline
\end{tabular}}
\caption{Adversarial evaluation after finetuning the baseline with adversarial policies derived from BayesAugment for generalization from SQuAD2.0 to TriviaQA, NewsQA, and SQuAD1.1 to German (de), Russian (ru) and Turkish (tr) RC datasets. Results (F1 / Exact Match) are shown on validation sets. Compare to corresponding rows in Table~\ref{tab:afterAdvTraining} to observe difference in performance between models finetuned with uniformly sampled dataset vs. dataset derived from learned policies.\label{tab:BayesAdv} \vspace{-15pt}}
\end{center}
\end{table*}

\textbf{Adversarial evaluation: } We show results from the adversarial evaluation of RoBERTa\textsubscript{BASE} models finetuned with adversarially augmented SQuAD using policies learned from BayesAugment in Table \ref{tab:BayesAdv}. We use the best models for out-of-domain and cross-lingual generalization as shown in Tables \ref{tab:mainDev} and \ref{tab:multilingualDev}, and evaluate their performance on the adversaries discussed in Section~\ref{sec:adveval}. Results show that the policies learnt from BayesAugment significantly improve resilience to the proposed adversarial attacks in addition to improving performance on the target datasets. The performance on adversaries varies with the transformation probability of the respective adversaries in the learned policies. For example, the transformation probability of \textit{PerturbQuestion} adversaries is 0.83 and 0.0 for SQuAD$\rightarrow$TriviaQA and SQuaD$\rightarrow$NewsQA models respectively (see Table~\ref{tab:sup_policies}). Consequently, the former has a higher performance on \textit{PerturbQuestion} adversaries.

\section{Analysis and Discussion}
Having established the efficacy of automated policy search for adversarial training, we further probe the robustness of adversarially trained models to unseen adversaries. We also analyze the convergence of BayesAugment for augmentation policy search and contrast its requirement of computational resources with that of AutoAugment. See Appendix for more analysis on domain independence of adversarial robustness and augmentation data size.

\textbf{Robustness to Unseen Adversaries:} \label{sec:UnseenAdv} 
We train RoBERTa\textsubscript{BASE} on SQuAD v2.0 augmented with the AddSentDiverse counterpart of each QA sample and evaluate it on other adversarial attacks, to analyze robustness of the model to unseen adversaries. As seen from the results in Table~\ref{tab:unseenAdv}, training with AddSentDiverse leads to large improvements on AddKSentDiverse and small improvements on PerturbQuestion and PerturbAnswer i.e., 31.21\% (45.31 vs. 76.52), 1.56\% (43.67 vs. 45.23) and 5.31\% (71.97 vs. 77.28) respectively, showing that the model becomes robust to multiple distractors within the same context and it also gains some resilience to paraphrasing operations. Conversely, we see a drop in performance on InvalidateAnswer, showing that it is easier for the model to be distracted by adversaries when the original answer is removed from context. 

\begin{table}[t]
\small
\begin{center}
\def\arraystretch{1.3}
\resizebox{0.47\textwidth}{!}{
\begin{tabular}{|l|l|l|} 
 \hline 
 {} & Trained on & Trained on \\
  Adversary Attack & SQuAD & SQuAD+AddSentDiverse \\
 \hline
 AddKSentDiverse & 45.31 & 76.52 \\ 
 InvalidateAnswer & 77.75 & 70.91 \\ 
 PerturbQuestion & 43.67 & 45.23 \\
 PerturbAnswer & 71.97 & 77.28 \\
 \hline 
\end{tabular}}
\vspace{-5pt}
\caption{Robustness of RoBERTa\textsubscript{BASE} trained on a subset of adversaries to unseen adversaries. Results (F1 score) are shown on SQuAD dev set.  \label{tab:unseenAdv}\vspace{-20pt}}
\end{center}
\end{table}

\begin{figure}[t]
\centering
\resizebox{0.4\textwidth}{!}{
    \includegraphics[width=0.5\textwidth]{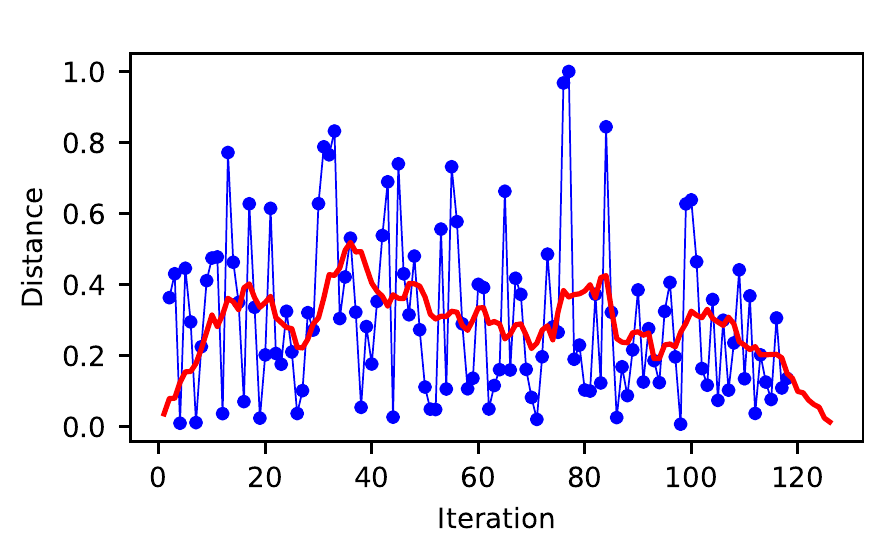}
    }
    \vspace{-7pt}
    \caption{Demonstration of variation in distance between neighboring samples picked by Bayesian optimizer with increasing training iterations. The red line represents moving average of distances. \label{fig:bayesianConvergence}\vspace{-11pt}}
\end{figure}

\textbf{Bayesian Convergence:}
In comparison to the thousands of training loops or more for AutoAugment, we run BayesAugment for only 100 training loops with 20 restarts. To show that BayesAugment converges within the given period, we plot the distance between transformation probabilities chosen by the Bayesian optimizer for the AddSentDiverse-PerturbQuestion augmentation method. As shown in Figure \ref{fig:bayesianConvergence}, the distance between the samples decreases with progression in training, showing that the optimizer becomes more confident about the narrow range of probability which should be sampled for maximum performance on validation set.

\textbf{Analysis of Resources for AutoAugment vs BayesAugment:}
With lesser number of training loops, BayesAugment uses only 10\% of the GPU resources required for AutoAugment. Our AutoAugment experiments have taken more than 1000 iterations and upto 5-6 days for convergence, requiring many additional days for hyperparameter tuning. In contrast, our BayesAugment experiment ran for 36-48 hours on 2 1080Ti GPUs and achieved comparable performance with 100 iterations or less. If large pretrained models are replaced with smaller distilled models in future work, BayesAugment will provide even more gains in time/computation.

\section{Conclusion}
We show that adversarial training can be leveraged to improve robustness of reading comprehension models to adversarial attacks and also to improve performance on source domain and generalization to out-of-domain and cross-lingual data. We present BayesAugment for policy search, which achieves results similar to the computationally-intensive AutoAugment method but with a fraction of computational resources. By combining policy search with rewards from the corresponding target development sets' performance, we show that models trained on SQuAD can be generalized to NewsQA and German, Russian, Turkish cross-lingual datasets without any training data from the target domain or language.

\section*{Acknowledgments}
We thank the reviewers for their useful feedback. This work was supported by DARPA MCS
Grant \#N66001-19-2-4031, DARPA KAIROS Grant \#FA8750-19-2-1004, ONR Grant \#N00014-18-1-2871, and awards from Google, Facebook, and Amazon (plus Amazon and Google GPU cloud credits). The views are those of the authors and not of the funding agency.

\bibliography{emnlp2020}
\bibliographystyle{acl_natbib}

\appendix

\section*{Appendix}
\section{Adversary Transformations}

We present two types of adversaries, namely positive perturbations and negative perturbations (or attacks). Positive perturbations are adversaries generated using methods that have been traditionally used for data augmentation in NLP i.e., semantic and syntactic transformations. Negative perturbations are adversaries based on the classic AddSent model \cite{jia2017adversarial} that exploit the RC model's shallow language understanding to mislead it to incorrect answers.

\textbf{AddSentDiverse:} We use the method outlined by \citet{wang2018robust} for AddSentDiverse to generate a distractor sentence and insert it randomly within the context of a QA sample. In addition to WordNet, we use ConceptNet \cite{speer2017conceptnet} for a wider choice of antonyms during generation of adversary. QA pairs that do not have an answer within the given context are also augmented with AddSentDiverse adversaries.

\textbf{AddKSentDiverse:} The AddSentDiverse method is used to generate multiple distractor sentences for a given context. Each of the distractor sentences is then inserted at independently sampled random positions within the context. The distractors may or may not be similar to each other. Introducing multiple points of confusion is a more effective technique for misleading the model and reduces the scope of learnable biases during adversarial training by adding variance.

\textbf{AddAnswerPosition:} The original answer span is retained and placed within a distractor sentence generated using a combination of AddSentDiverse and random perturbations to maximize semantic mismatch. We modify the evaluation script to compare exact answer span locations in addition to the answer phrase and fully penalize incorrect locations. For practical purposes, if the model predicts the answer span within adversarial sentence as output, it does not make a difference. However, it brings into question the interpretability of such models. This distractor is most effective when placed right before the original answer sentence, showing dependence on insert location of distractor.

\textbf{InvalidateAnswer:} The sentence containing the original answer is removed from the context. Instead, a distractor sentence generated using AddSentDiverse is introduced to the context. This method is used to augment the adversarial \textit{NoAnswer}-style samples in SQuAD v2.0.

\textbf{PerturbAnswer (Semantic Paraphrasing):}
Following \citet{alzantot2018generating}, we perform semantic paraphrasing of the sentence containing the answer span. Instead of using genetic algorithm, we adapt their \texttt{Perturb} subroutine to generate paraphrases in the following steps: 
\begin{enumerate}[nosep, wide=0pt, leftmargin=*, after=\strut]
    \item Select word locations for perturbations, which includes locations within any content phrase that does not appear within the answer span. Here, content phrases are verbs, adverbs and adjectives.
    \item For location $k_{i}$ in the set of word locations $\{k\}$, compute 20 nearest neighbors of the word at given location using GloVe embeddings, create a candidate sentence by perturbing the word location with each of the substitute words and rank perturbed sentences using a language model.
    \item Select the perturbed sentence with highest rank and perform Step 2 for the next location $k_{i+1}$ using the perturbed sentence.
\end{enumerate}
We use the OpenAI-GPT model \cite{radford2018improving} to evaluate paraphrases.

\textbf{PerturbQuestion (Syntactic Paraphrasing):}
We use the syntactic paraphrase network introduced by \citet{iyyer2018adversarial} to generate syntactic adversaries. Sentences from the context of QA samples tend to be long and have complicated syntax. The corresponding syntactic paraphrases generated by the paraphrasing network usually miss out on half of the source sentence. Therefore, we choose to perform paraphrasing on the questions. We generate 10 paraphrases for each question and rank them based on cosine similarity, computed between the mean of word embeddings \cite{pennington2014glove} of source sentence and generated paraphrases \cite{niu2018adversarial, liu2016not}.

Finally, we combine negative perturbations with positive perturbations to create adversaries which double-down on the model's language understanding capabilities. It always leads to a larger drop in performance when tested on the reading comprehension models trained on original unaugmented datasets.

\begin{table*}[t]
\small
\begin{center}
\def\arraystretch{1.3}
\resizebox{0.9\textwidth}{!}{
\begin{tabular}{|ll|}
 \hline
 \multicolumn{2}{|c|}{AutoAugment Policies} \\
 \hline
 SQuAD $\rightarrow$ SQuAD &  (AddS, None, 0.2) $\rightarrow$ (IA, None, 0.4) $\rightarrow$ (AddA, None, 0.2)\\
 SQuAD $\rightarrow$ NewsQA &  (None, PA, 0.4) $\rightarrow$ (None, PA, 0.6) $\rightarrow$ (AddS, PA, 0.4)\\ 
  SQuAD $\rightarrow$ TriviaQA &  (AddS, None, 0.9) $\rightarrow$ (AddS, PA, 0.7) $\rightarrow$ (AddKS, PQ, 0.9)\\ 
 NewsQA $\rightarrow$ NewsQA & (AddA, PA, 0.2) $\rightarrow$ (AddKS, None, 0.2) $\rightarrow$ (AddA, PA, 0.4) \\
 \hline
  \multicolumn{2}{|c|}{BayesAugment Policies} \\
 \hline
 SQuAD $\rightarrow$ SQuAD &  (AddS, 0.29), (AddA, 0.0), (AddA-PA, 0.0), (AddA-PQ, 0.0), (AddKS, 0.0), (AddKS-PA,0.0)\\
 & (AddKS-PQ, 0.0), (AddS-PA, 0.0), (AddS-PQ, 0.0), (PA, 0.61), (PQ, 0.0), (IA, 1.0)\\ 
 & \\
 SQuAD $\rightarrow$ NewsQA & (AddS, 1.0), (AddA, 0.0), (AddA-PA, 1.0), (AddA-PQ, 0.0), (AddKS, 0.0), (AddKS-PA, 0.0)\\
 & (AddKS-PQ, 0.0), (AddS-PA, 1.0), (AddS-PQ, 0.0), (PA, 0.48), (PQ, 0.0), (IA, 0.0)\\ 
 & \\
  SQuAD $\rightarrow$ TriviaQA & (AddS, 1.0), (AddA, 1.0), (AddA-PA, 0.21), (AddA-PQ, 0.18), (AddKS, 0.86), (AddKS-PA, 0.37)\\
 & (AddKS-PQ, 0.25), (AddS-PA, 0.12), (AddS-PQ, 0.49), (PA, 0.91), (PQ, 0.83), (IA, 0.26)\\ 
 & \\
 SQuAD $\rightarrow$ MLQA(de) & (AddS, 0.042), (AddA-PA, 0.174), (AddA-PQ, 0.565), (AddKS, 0.173), (AddKS-PA, 0.567)\\
 & (AddA, 0.514), (AddS-PA, 0.869), (AddS-PQ, 0.720), (PA, 0.903), (PQ, 0.278), (AddKS-PQ, 0.219)\\ 
 & \\
 SQuAD $\rightarrow$ XQuAD(ru) &  (AddS, 0.147), (AddA-PA, 0.174), (AddA-PQ, 0.79), (AddKS, 0.55), (AddKS-PA, 0.97)\\
 & (AddA, 0.77), (AddS-PA, 0.02), (AddS-PQ, 0.59), (PA, 0.11), (PQ, 0.95), (AddKS-PQ, 0.725)\\
 & \\
 SQuAD $\rightarrow$ XQuAD(tr) &  (AddS, 0.091), (AddA-PA, 0.463), (AddA-PQ, 0.64), (AddKS, 0.32), (AddKS-PA, 0.86)\\
 & (AddA, 0.34), (AddS-PA, 0.37), (AddS-PQ, 0.43), (PA, 0.27), (PQ, 0.81), (AddKS-PQ, 0.493)\\
 & \\
 NewsQA $\rightarrow$ NewsQA & (AddS, 1.0), (AddA, 1.0), (AddA-PA, 1.0), (AddA-PQ, 0.0), (AddKS, 0.0), (AddKS-PA, 1.0)\\
 & (AddKS-PQ, 0.156), (AddS-PA, 0.0), (AddS-PQ, 0.720), (PA, 0.0), (PQ, 0.0), (IA, 1.0) \\
 \hline
\end{tabular}}
\caption{Best Policies suggested by BayesAugment and AutoAugment methods for different scenarios; AddS = AddSentDiverse, AddKS = AddKSentDiverse, AddA = AddAnswerPosition, IA = InvalidateAnswer, PA = PerturbAnswer, PQ = PerturbQuestion. \label{tab:sup_policies}}
\end{center}
\end{table*}

\textbf{Semantic Difference Check:}
To make sure that the distractor sentences are sufficiently different from the original sentence, we perform a semantic difference check in two steps:
\begin{enumerate}[nosep, wide=0pt, leftmargin=*, after=\strut]
    \item Extract content phrases from original sentence. Content phrase is any common NER phrase or one of the four: noun, verb, adverb, adjective.
    \item There should be at least 2 content phrases in the original text that aren't found in the distractor.
\end{enumerate}
We examined 100 randomly sampled original-distractor sentence pairs and found that our semantic difference check works for 96\% of the cases.

\section{BayesAugment}
We use Gaussian Process (GP) \cite{rasmussen2003gaussian} as surrogate function and Upper Confidence Bound (UCB) \cite{srinivas2009gaussian} as the acquisition function. GP is a non-parametric model that is fully characterized by a mean function $\mu_{0}: \chi \mapsto {\rm I\!R}$ and a positive-definite kernel or covariance function $k: \chi \times \chi \mapsto {\rm I\!R}$. Let $x_{1}, x_{2},... x_{n}$ denote any finite collections of $n$ points, where each $x_{i}$ represents a choice of sampling probabilities for each of the augmentation methods and $f_{i} = f(x_{i})$ is the (unknown) function value evaluated at $x_{i}$. Let $y_{1}, y_{2},... y_{n}$ be the corresponding noisy observations (the validation performance at the end of training). In the context of GP Regression (GPR), $f={f_{1}, ..... f_{n}}$ are assumed to be jointly Gaussian. Then, the noisy observations $y = y_{1}, .... y_{n}$ are normally distributed around $f$ as $y|f \sim \mathcal{N} (f, \sigma^{2}I)$. The Gaussian Process upper confidence bound (GP-UCB) algorithm measures the optimistic performance upper bound of the sampling probabilities.

\section{Datasets}

\textbf{SQuAD v2.0} \cite{rajpurkar2018know} is a crowd-sourced dataset consisting of 100,000 questions from SQuAD v1.1 \cite{rajpurkar2016squad} and an additional 50,000 questions that do not have answers within the given context. We split the official development set into 2 randomly sampled sets of validation and test for our experiments.

\textbf{NewsQA} is also a crowd-sourced extractive RC dataset based on 10,000 news articles from CNN, containing both answerable and unanswerable questions. \cite{trischler2016newsqa} To accommodate very long contexts from NewsQA in models like Bert \cite{devlin2018bert} and RoBERTa \cite{liu2019roberta}, we sample two instances from the set of overlapping instances for the final training data.

\textbf{TriviaQA} \cite{joshi2017triviaqa} questions were crawled from the web and have two variants. One variant includes Wikipedia articles as contexts; we use the other variant which involves web snippets and documents from Bing search engine as contexts. The development and test sets are large with more than 60K samples in each. For faster BayesAugment and AutoAugment iterations, we randomly select 10K samples from the development set to generate rewards.

\textbf{MLQA} \cite{lewis2019mlqa} is the multilingual extension to SQuAD v1.1 consisting of evaluation (development and test) data only. We use German (de) MLQA in our experiments.

\textbf{XQuAD} is a multilingual version of SQuAD \cite{artetxe2019cross} containing only test sets. We use Russian (ru) and Turkish (tr) XQuAD which contain nearly 1100 QA samples that are further split equally and randomly into development and test sets.

\begin{table}[t]
\small
\begin{center}
\def\arraystretch{1.3}
\resizebox{0.43\textwidth}{!}{
\makebox[0.5\textwidth]{
\begin{tabular}{|p{19mm}p{17mm}p{17mm}p{17mm}|}
 \hline
 Model & SQuADv1.1 & SQuADv2.0 & NewsQA \\
 \hline
 RoBERTa & 89.73 / 82.38 & 81.17 / 77.54 & 58.40 / 47.04 \\
 DistilRoBERTa & 84.57 / 75.81 & 73.29 / 69.47 & 54.21 / 42.76 \\ 
 \hline
\end{tabular}}}
\caption{Comparison of performance (F1 Score / Exact Match) of different models on SQuAD v1.1, SQuaD v2.0 and NewsQA datasets. RoBERTa\textsubscript{BASE} is the baseline model; DistilRoBERTa\textsubscript{BASE} is the task model used during AutoAugment policy search. \label{tab:baseline}}
\end{center}
\end{table}

\begin{table}[t]
\small
\begin{center}
\def\arraystretch{1.3}
\resizebox{0.47\textwidth}{!}{
\begin{tabular}{|p{25mm}|p{17mm}|p{17mm}|p{13mm}|} 
 \hline
 \textbf{Hyperparameter} & \textbf{SQuAD v1.1} & \textbf{SQuAD v2.0} & \textbf{NewsQA} \\
 \hline
 Learning Rate & 3e-5 & 1.5e-5 & 1.6e-5 \\
 Batch Size & 24 & 16 & 24 \\ 
 Warmup Ratio & 0.06 & 0.06 & 0.08 \\ 
 No. of Epochs & 2 & 5 & 5 \\
 Weight Decay & 0.01 & 0.01 & 0.01\\
 \hline
\end{tabular}}
\caption{Best hyperparameters for training RoBERTa\textsubscript{BASE} on SQuAD v2.0 and NewsQA.\label{tab:hyperparams}}
\end{center}
\end{table}

\section{Training Details}

\textbf{Reading Comprehension Models:} We use RoBERTa\textsubscript{BASE} as the primary RC model for all our experiments. Search algorithms like AutoAugment require a downstream model that can be trained and evaluated fast, in order to reduce training time. So, we use distilRoBERTa\textsubscript{BASE} \cite{sanh2019distilbert} for AutoAugment training loops, which has 40\% lesser parameters than RoBERTa\textsubscript{BASE}. It should be noted that the distilRoBERTa model used in our experiments is trained on SQuAD without distillation. BayesAugment is trained for fewer iterations than AutoAugment and hence, allows us to use RoBERTa\textsubscript{BASE} model directly in the training loop.

\begin{table}[t]
\small
\begin{center}
\def\arraystretch{1.3}
\begin{tabular}{|p{30mm}|p{30mm}|} 
 \hline
 \textbf{Hyperparameter} & \textbf{Range} \\
 \hline
 Learning Rate & $[1e^{-5}, 2e^{-5}]$\\
Batch Size & $\{8, 16, 24, 32\}$ \\
Warmup Ratio & $[0.01, 0.5]$ \\
Weight Decay & $[0.01, 0.1]$\\
 \hline
\end{tabular}
\vspace{-5pt}
\caption{Bayesian Optimization Ranges for Finetuning RoBERTA with AutoAugment and Bayesaugment policies (32 iterations with 8 restarts). \label{tab:bayesopt_params}\vspace{-10pt}}
\end{center}
\end{table}

\textbf{Model Hyperparameters:} We trained RoBERTa\textsubscript{BASE} for 5 epochs on SQuAD and NewsQA respectively and selected the best-performing checkpoint as baseline. We perform a hyperparameter search for both datasets using Bayesian optimization search \cite{snoek2012practical}. The RNN controller in AutoAugment training loop consists of a single LSTM cell with a single hidden layer and hidden layer dimension of 100. The generated policy consists of 3 sub-policies; each sub-policy is structured as discussed in main text. BayesAugment is trained for 100 iterations with 20 restarts. During AutoAugment and BayesAugment training loops, RoBERTa\textsubscript{BASE} or distilRoBERTa\textsubscript{BASE} (which has already been trained on unaugmented SQuAD) is further finetuned on the adversarially augmented dataset for 2 epochs with a warmup ratio of 0.2 and learning rate decay (lr=1e-5) thereafter. After the policy search, further hyperparameter optimization is performed for best results from fine-tuning. We do not perform this last step of hyperparameter tuning on cross-lingual data to avoid the risk of overfitting the small datasets. For generalization from SQuAD v1.1 to cross-lingual datasets, we do not consider the adversary InvalidateAnswer because \textit{NoAnswer} samples do not exist for these datasets. Refer to Tables~\ref{tab:hyperparams} and \ref{tab:bayesopt_params} for more details on hyperparameters.

\section{Analysis}
In this section, we show the impact of adversarial augmentation ratio in training dataset and the size of training dataset on the generalization of RC model to out-of-domain data. Next, we show more experiments on robustness to unseen adversaries. Finally, we analyze the domain-independence of adversarial robustness by training on adversarially augmented SQuAD and testing on adversarial NewsQA samples.

\begin{table}[t]
\small
\begin{center}
\def\arraystretch{1.3}
\resizebox{0.45\textwidth}{!}{
\begin{tabular}{|l|ll|}
 \hline
 NewsQA Adversary & SQuAD & SQuAD $\rightarrow$ \\
 {} & {} & {NewsQA}\\
 \hline
 AddSentDiverse & 42.39 / 32.79 & 49.54 / 38.02 \\
PerturbAnswer & 39.95 / 27.60 & 45.52 / 32.49 \\
AddSentDiv-PertrbAns & {35.08 / 26.33} & {43.63 / 32.76} \\
 \hline
\end{tabular}}
\vspace{-5pt}
\caption{Comparison of robustness between RoBERTa\textsubscript{BASE} finetuned on original unaugmented SQuAD and our best SQuAD $\rightarrow$ NewsQA generalized model. Results (F1 score/Exact Match) are shown on dev set.\label{tab:robustnessTransfer}\vspace{-15pt}}
\end{center}
\end{table}

\begin{table}[t!]
\small
\begin{center}
\def\arraystretch{1.3}
\resizebox{0.39\textwidth}{!}{
\begin{tabular}{|p{40mm}|p{20mm}|} 
 \hline
 \textbf{Augmentation Ratio} & \textbf{NewsQA} \\
 \hline
 RoBERTa & 48.36 / 36.06 \\
 \hspace{1cm} + 1x augmentation & 49.73 / 38.38 \\ 
 \hspace{1cm} + 2x augmentation & 49.84 / 37.97 \\ 
 \hspace{1cm} + 3x augmentation & 49.62 / 38.01 \\ 
 \hline
\end{tabular}}
\caption{Effect of augmentation ratio for generalization from SQuAD$\rightarrow$NewsQA. Results (F1 score/Exact Match) are shown on NewsQA dev set.\label{tab:augRatio}}
\end{center}
\end{table}

\begin{figure}[t]
\centering
    \includegraphics[width=0.45\textwidth]{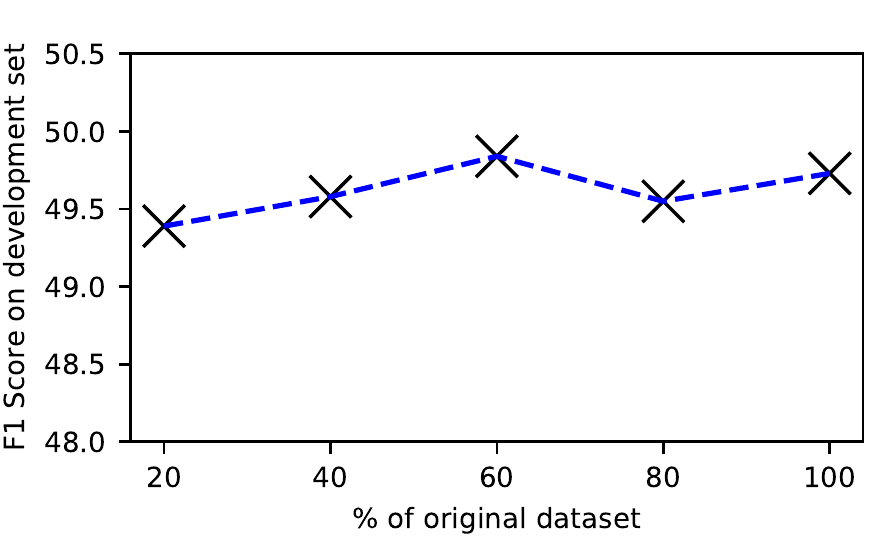}
    \caption{Performance of SQuAD $\rightarrow$ NewsQA model on NewsQA dev set (F1 score) with increasing size of finetuning dataset. \label{fig:augmentationsize}}
\end{figure}

\textbf{Effect of Augmentation Ratio:} To assess the importance of adversarial augmentation in the dataset, we experimented with different ratios i.e., 1x, 2x and 3x, of augmented samples to the original dataset, for generalization from SQuAD to NewsQA using the augmentation policy learnt by BayesAugment. The performance of SQuAD$\rightarrow$NewsQA models on NewsQA validation set were 49.73, 49.84 and 49.62 for 1x, 2x and 3x augmentations respectively, showing slight improvement for twice the number of augmentations (see Table~\ref{tab:augRatio}). However, the performance starts decreasing at 3x augmentations, showing that too many adversaries in the training data starts hurting generalization.

\textbf{Effect of Augmented Dataset Size:} We experimented with 20\%, 40\%, 60\%, 80\% and 100\% of the original dataset to generate augmented dataset using the BayesAugment policy for generalization of RoBERTa\textsubscript{BASE} trained on SQuAD to NewsQA and observed little variance in performance with increasing data, as seen from Figure~\ref{fig:augmentationsize}. The augmentation ratio in these datasets is 1:1. We hypothesize that the model is saturated early on during training, within the first tens of thousands of adversarially augmented samples. Exposing the model to more SQuAD samples gives little boost to performance on NewsQA thereafter.

\begin{table}[t]
\small
\begin{center}
\def\arraystretch{1.3}
\resizebox{0.47\textwidth}{!}{
\begin{tabular}{|p{45mm}|p{15mm}|p{20mm}|} 
 \hline
  & \textbf{Trained on}  & \textbf{Trained on} \\
  \textbf{Adversary Attack} & \textbf{SQuAD} & \textbf{SQ+ASD/PQ/PA} \\
 \hline
AddSentDiverse+PerturbAnswer & 50.71 & 84.37 \\
AddKSentDiverse+PerturbQuestion & 31.56 & 78.91 \\
AddAnswerPosition & 68.91 & 80.87 \\
AddKSentDiverse & 45.31 & 76.14 \\
InvalidateAnswer & 77.75 & 71.62 \\
 \hline
\end{tabular}}
\caption{Robustness of RoBERTa\textsubscript{BASE} trained on a subset of adversaries to unseen adversaries. Results (F1 score) are shown on SQuAD dev set (ASD=AddSentDiverse, PQ=PerturbQuestion, PA=PerturbAnswer, SQ=SQuAD).  \label{tab:unseenAdvSup}}
\end{center}
\end{table}

 \textbf{Robustness to Unseen Adversaries:} We train RoBERTa\textsubscript{BASE} on SQuAD which has been augmented with an adversarial dataset of the same size as SQuAD and contains equal number of samples from AddSentDiverse, PerturbQuestion and PerturbAnswer. In Table \ref{tab:unseenAdvSup}, We see that the model is significantly more robust to combinatorial adversaries like AddSentDiverse+PerturbAnswer when trained on the adversaries AddSentDiverse and PerturbAnswer individually. We also see a decline in performance on InvalidateAnswer.
 
 \textbf{Domain-Independence of Robustness to Adversarial Attacks:}
We have shown that a reading comprehension model trained on SQuAD can be generalized to NewsQA by finetuning the model with adversarially transformed samples from SQuAD dataset. It is expected that this model will be robust to similar attacks on SQuAD. To assess if this robustness generalizes to NewsQA as well, we evaluate our best SQuAD$\rightarrow$NewsQA model on adversarially transformed NewsQA samples from the development set. The SQuAD column in Table~\ref{tab:robustnessTransfer} shows results from evaluation of RoBERTa\textsubscript{BASE} finetuned with original unaugmented SQuAD, on adversarially transformed NewsQA samples. Interestingly, the generalized model (rightmost column) is 5-8\% more robust to adversarial NewsQA without being trained on any NewsQA samples, showing that robustness to adversarial attacks in source domain easily generalizes to a different domain.

\end{document}